\documentclass{article} 
\usepackage{iclr2024_conference_style_modified,times}

\usepackage{amsmath,amsfonts,bm}

\def\eqref#1{equation~\ref{#1}}

\def\1{\bm{1}}

\DeclareMathAlphabet{\mathsfit}{\encodingdefault}{\sfdefault}{m}{sl}
\SetMathAlphabet{\mathsfit}{bold}{\encodingdefault}{\sfdefault}{bx}{n}

\newcommand{\R}{\mathbb{R}}

\usepackage{extra_preamble}

\usepackage{hyperref}

\title{\textbf{Sparse Autoencoders Find Highly Interpretable Features in Language Models}}

\author{Hoagy Cunningham\thanks{Equal contribution}\hspace{4pt}$^{1 2}$,\  Aidan Ewart$^{* 1 3}$, \ Logan Riggs$^{* 1}$, \ Robert Huben, \ Lee Sharkey$^{4}$ \\
$^1$EleutherAI, $^2$MATS, $^3$Bristol AI Safety Centre, $^4$Apollo Research \\
\texttt{\{hoagycunningham, aidanprattewart, logansmith5\}@gmail.com}}

\noniclrversion 
\begin{document}

\maketitle

\begin{abstract}

One of the roadblocks to a better understanding of neural networks' internals is \textit{polysemanticity}, where neurons appear to activate in multiple, semantically distinct contexts. Polysemanticity prevents us from identifying concise, human-understandable explanations for what neural networks are doing internally. One hypothesised cause of polysemanticity is \textit{superposition}, where neural networks represent more features than they have neurons by assigning features to an overcomplete set of directions in activation space, rather than to individual neurons. Here, we attempt to identify those directions, using sparse autoencoders to reconstruct the internal activations of a language model. These autoencoders learn sets of sparsely activating features that are more interpretable and monosemantic than directions identified by alternative approaches, where interpretability is measured by automated methods. Moreover, we show that with our learned set of features, we can pinpoint the features that are causally responsible for counterfactual behaviour on the indirect object identification task \citep{wang2022interpretability} to a finer degree than previous decompositions. This work indicates that it is possible to resolve superposition in language models using a scalable, unsupervised method. Our method may serve as a foundation for future mechanistic interpretability work, which we hope will enable greater model transparency and steerability.

\end{abstract}

\section{Introduction}
Advances in artificial intelligence (AI) have resulted in the development of highly capable AI systems that make decisions for reasons we do not understand. This has caused concern that AI systems that we cannot trust are being widely deployed in the economy and in our lives, introducing a number of novel risks \citep{hendrycks2023overview}, including potential future risks that AIs might deceive humans in order to accomplish undesirable goals \citep{ngo2022alignment}. Mechanistic interpretability seeks to mitigate such risks through understanding how neural networks calculate their outputs, allowing us to reverse engineer parts of their internal processes and make targeted changes to them \citep{cammarata2021curve, wang2022interpretability, elhage2021mathematical}.

\ificlrfinal
\blfootnote{Code to replicate experiments can be found at \url{https://github.com/HoagyC/sparse_coding}}
\fi

To reverse engineer a neural network, it is necessary to break it down into smaller units (features) that can be analysed in isolation. Using individual neurons as these units has had some success \citep{olah2020zoom, bills2023language}, but a key challenge has been that neurons are often \textit{polysemantic}, activating for several unrelated types of feature \citep{olah2020zoom}. Also, for some types of network activations, such as the residual stream of a transformer, there is little reason to expect features to align with the neuron basis \citep{elhage2023privileged}.

\citet{elhage2022toy} investigate why polysemanticity might arise and hypothesise that it may result from models learning more distinct features than there are dimensions in the layer. They call this phenomenon \textit{superposition}. Since a vector space can only have as many orthogonal vectors as it has dimensions, this means the network would learn an overcomplete basis of non-orthogonal features. Features must be sufficiently sparsely activating for superposition to arise because, without high sparsity, interference between non-orthogonal features prevents any performance gain from superposition. This suggests that we may be able to recover the network's features by finding a set of directions in activation space such that each activation vector can be reconstructed from a sparse linear combinations of these directions. This is equivalent to the well-known problem of sparse dictionary learning  \citep{olshausen1997sparse}. 

\begin{figure}
    \centering
    \includegraphics[width=1\textwidth]{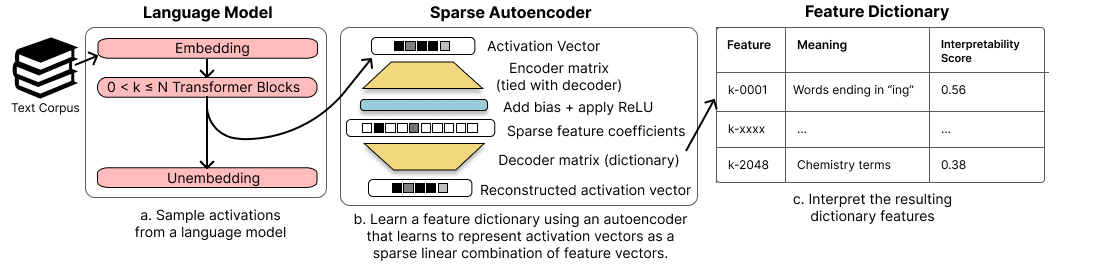}
    \caption{An overview of our method. We a) sample the internal activations of a language model, either the residual stream, MLP sublayer, or attention head sublayer; b) use these activations to train a neural network, a sparse autoencoder whose weights form a feature dictionary; c) interpret the resulting features with techniques such as OpenAI's autointerpretability scores.}
    \label{fig:overview}
\end{figure}

Building on \citet{sharkey2023technical}, we train sparse autoencoders to learn these sets of directions. Our approach is also similar to \citet{yun2021transformer}, who apply sparse dictionary learning to all residual stream layers in a language model simultaneously. Our method is summarised in Figure \ref{fig:overview} and described in Section \ref{section:dictionary_learning}.

We then use several techniques to verify that our learned features represent a semantically meaningful decomposition of the activation space. First, we show that our features are on average more interpretable than neurons and other matrix decomposition techniques, as measured by autointerpretability scores (Section \ref{section:autointerp}) \citep{bills2023language}. Next, we show that we are able to pinpoint the features used for a set task more precisely than other methods (Section \ref{section:identifying_causal_features}). Finally, we run case studies on a small number of features, showing that they are not only monosemantic but also have predictable effects on the model outputs, and can be used for fine-grained circuit detection. (Section \ref{section:monosemantic}).

\section{Taking Features out of Superposition with Sparse Dictionary Learning}
\label{section:dictionary_learning}
\label{section:sparse_autoencoder}
To take network features out of superposition, we employ techniques from \emph{sparse dictionary learning} \citep{olshausen1997sparse, lee2006efficient}. Suppose that each of a given set of vectors $\{\mathbf{x}_i\}_{i=1}^{n_{\text{vec}}} \subset \R^d$ is composed of a sparse linear combination of unknown vectors $\{\mathbf{g}_j\}_{j=1}^{n_{\text{gt}}} \subset \R^d$, i.e. $\mathbf{x}_i = \sum_j a_{i,j}\mathbf{g}_j$ where $\mathbf{a_i}$ is a sparse vector. In our case, the data vectors $\{\mathbf{x}_i\}_{i=1}^{n_{\text{vec}}}$ are internal activations of a language model, such as Pythia-70M \citep{biderman2023pythia}, and $\{\mathbf{g}_j\}_{j=1}^{n_{\text{gt}}}$ are unknown, ground truth network features. We would like learn a dictionary of vectors, called dictionary features, $\{\mathbf{f}_k\}_{k=1}^{n_{\text{feat}}} \subset \R^d$ where for any network feature $\mathbf{g}_j$ there exists a dictionary feature $\mathbf{f}_k$ such that $\mathbf{g}_j \approx \mathbf{f}_k$. 

To learn the dictionary, we train an autoencoder with a sparsity penalty term on its hidden activations. The autoencoder is a neural network with a single hidden layer of size $d_{\text{hid}}=R d_{\text{in}}$, where $d_{\text{in}}$ is the dimension of the language model internal activation vectors\footnote{We mainly study residual streams in Pythia-70M and Pythia 410-M, for which the residual streams are of size $d_{\text{in}}=512$ and $d_{\text{in}}=1024$, respectively \citep{biderman2023pythia}}, and $R$ is a hyperparameter that controls the ratio of the feature dictionary size to the model dimension. We use the ReLU activation function in the hidden layer \citep{fukushima1975relu}. We also use tied weights for our neural network, meaning the weight matrices of the encoder and decoder are transposes of each other.\footnote{We use tied weights because (a) they encode our expectation that the directions which detect and define the feature should be the same or highly similar, (b) they halve the memory cost of the model, and 
(c) they remove ambiguity about whether the learned direction should be interpreted as the encoder or decoder direction. They do not reduce performance when training on residual stream data but we have observed some reductions in performance when using MLP data.}  Thus, on input vector $\mathbf{x} \in \{\mathbf{x}_i\}$, our network produces the output $\mathbf{\hat x}$, given by
\begin{eqnarray}
\mathbf{c} &=& \mathrm{ReLU}(M\mathbf{x}+\mathbf{b}) \label{eq:encoder}\\
\mathbf{\hat x} &=& M^T \mathbf{c} \label{eq:decoder}\\
&=& \sum_{i = 0}^{d_{\text{hid}}-1} c_i \mathbf{f}_i
\end{eqnarray}

where $M \in \R^{d_{\text{hid}} \times d_{\text{in}}}$ and $\mathbf{b} \in \R^{d_{\text{hid}}}$ are our learned parameters, and $M$ is normalised row-wise\footnote{Normalisation of the rows (dictionary features) prevents the model from reducing the sparsity loss term $||\mathbf{c}||_1$ by increasing the size of the feature vectors in $M$.}. Our parameter matrix $M$ is our feature dictionary, consisting of $d_{\text{hid}}$ rows of dictionary features $\mathbf{f}_i$. The output $\mathbf{\hat x}$ is meant to be a reconstruction of the original vector $\mathbf x$, and the hidden layer $\mathbf c$ consists of the coefficients we use in our reconstruction of $\mathbf x$.



Our autoencoder is trained to minimise the loss function

\begin{equation}
\label{eqn:loss_function}
    \mathcal L(\mathbf x)= \underbrace{||\mathbf{x}- \mathbf{\hat x}||_2^2}_{\text{Reconstruction loss}} + \underbrace{\alpha ||\mathbf{c}||_1}_{\text{Sparsity loss}}
\end{equation}

where $\alpha$ is a hyperparameter controlling the sparsity of the reconstruction. The $\ell^1$ loss term on $\mathbf c$ encourages our reconstruction to be a sparse linear combination of the dictionary features. It can be shown empirically \citep{sharkey2023technical} and theoretically \citep{wright2022high} that reconstruction with an $\ell^1$ penalty can recover the ground-truth features that generated the data. For the further details of our training process, see Appendix \ref{appendix:hparam_selection}.

\section{Interpreting Dictionary Features}
\label{section:autointerp}
\subsection{Interpretability at Scale}
\label{section:autointerp_background}
Having learned a set of dictionary features, we want to understand whether our learned features display reduced polysemanticity, and are therefore more interpretable. To do this in a scalable manner, we require a metric to measure how interpretable a dictionary feature is. We use the automated approach introduced in \citet{bills2023language} because it scales well to measuring interpretability on the thousands of dictionary features our autoencoders learn. In summary, the autointerpretability procedure takes samples of text where the dictionary feature activates, asks a language model to write a human-readable interpretation of the dictionary feature, and then prompts the language model to use this description to predict the dictionary feature's activation on other samples of text. The correlation between the model's predicted activations and the actual activations is that feature's interpretability score. See Appendix \ref{appendix:autointerp_protocol} and \citet{bills2023language} for further details.

We show descriptions and top-and-random scores for five dictionary features from the layer 1 residual stream in Table \ref{table:autointerpretation}. The features shown are the first five under the (arbitrary) ordering in the dictionary.

\begin{table}
    \centering
\begin{tabular}{|c|p{8cm}|c|}  
 \hline
 Feature & Description (Generated by GPT-4) & Interpretability Score \\ 
 \hline
 1-0000 & parts of individual names, especially last names. & 0.33 \\ 
 \hline
 1-0001 & actions performed by a subject or object. & -0.11 \\
 \hline
 1-0002 & instances of the letter `W' and words beginning with `w'. & 0.55 \\ 
 \hline
 1-0003 & the number `5' and also records moderate to low activation for personal names and some nouns. & 0.57 \\
 \hline
 1-0004 & legal terms and court case references. & 0.19 \\
 \hline
\end{tabular}
    \caption{Results of autointerpretation on the first five features found in the layer 1 residual stream. Autointerpretation produces a description of what the feature means and a score for how well that description predicts other activations.}
    \label{table:autointerpretation}
\end{table}

\subsection{Sparse dictionary features are more interpretable than baselines}
\label{subsection:interpretability_baselines}
We assess our interpretability scores against a variety of alternative methods for finding dictionaries of features in language models. In particular, we compare interpretability scores on our dictionary features to those produced by a) the default basis, b) random directions, c) Principal Component Analysis (PCA), and d) Independent Component Analysis (ICA). For the random directions and for the default basis in the residual stream, we replace negative activations with zeros so that all feature activations are nonnegative \footnote{
For PCA we use an online estimation approach and run the decomposition on the same quantity of data we used for training the autoencoders. For ICA, due to the slower convergence times, we run on only 2GB of data, approximately 4 million activations for the residual stream and 1m activations for the MLPs.}.


\begin{figure}
    \centering
    \includegraphics[scale=0.6]{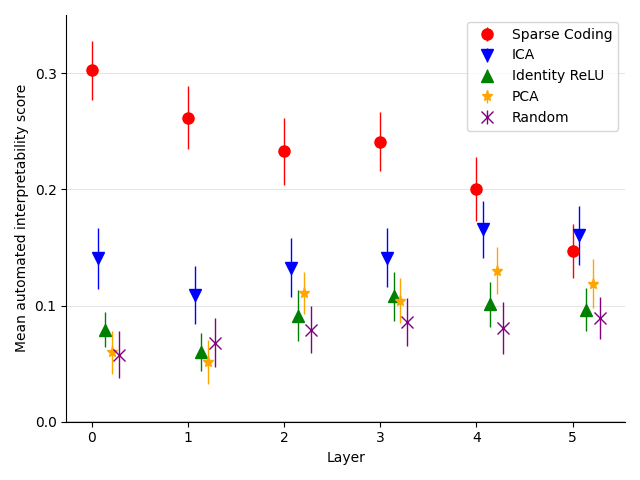}
    \caption{Average top-and-random autointerpretability score of  our learned directions in the residual stream, compared to a number of baselines, using 150 features each. Error bars show 95\% confidence intervals around means. The feature dictionaries used here were trained for 10 epochs using $\alpha=.00086$ and $R=2$.}
    \label{fig:main_interp_scores_comparison}
\end{figure}

Figure \ref{fig:main_interp_scores_comparison} shows that our dictionary features are far more interpretable by this measure than dictionary features found by comparable techniques. We find that the strength of this effect declines as we move through the model, being comparable to ICA in layer 4 and showing minimal improvement in the final layer.

This could indicate that sparse autoencoders work less well in later layers but also may be connected to the difficulties of automatic interpretation, both because by building on earlier layers, later features may be more complex, and because they are often best explained by their effect on the output. \citet{bills2023language} showed that GPT-4 is able to generate explanations that are very close to the average quality of the human-generated explanations given similar data. However, they also showed that current LLMs are limited in the kinds of patterns that they can find, sometimes struggling to find patterns that center around next or previous tokens rather than the current token, and in the current protocol are unable to verify outputs by looking at changes in output or other data. 

We do show, in Section \ref{section:case_studies}, a method to see a feature's causal effect on the output logits by hand, but we currently do not send this information to the language model for hypothesis generation. The case studies section also demonstrates a closing parenthesis dictionary feature, showing that these final layer features can give insight into the model's workings.

See Appendix \ref{appendix:full_autointerp} for a fuller exploration of different learned dictionaries through the lens of automatic interpretability, looking at both the MLPs and the residual stream.

\section{Identifying Causally-Important Dictionary Features for Indirect Object Identification}
\label{section:identifying_causal_features}
\begin{figure}[!b]
    \centering
    \includegraphics[width=\textwidth]{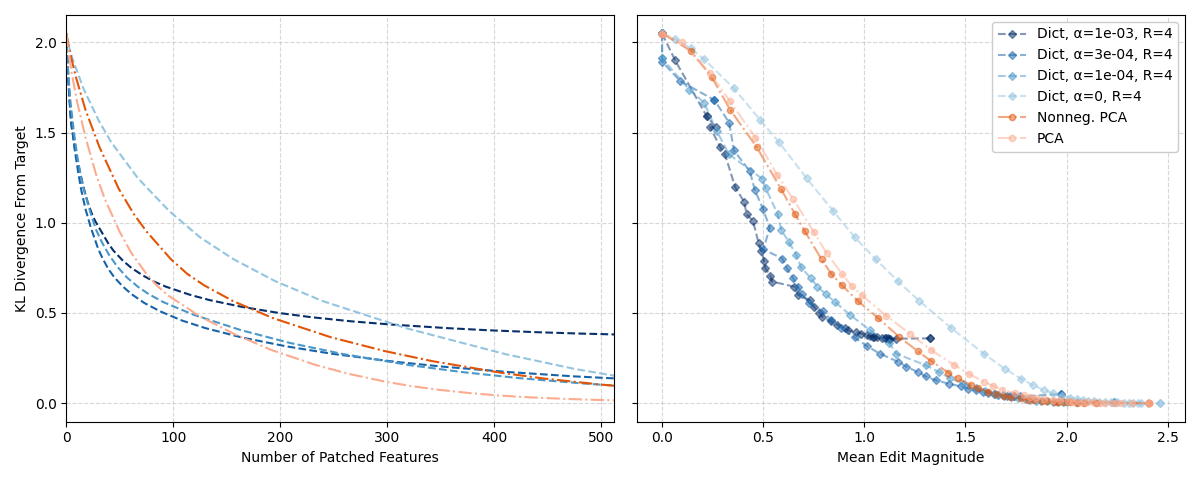}
    \caption{(Left) Number of features patched vs KL divergence from target, using various residual stream decompositions. We find that patching a relatively small number of dictionary features is more effective than patching PCA components and features from the non-sparse $\alpha=0$ dictionary. (Right) Mean edit magnitude vs KL divergence from target as we increase the number of patched features. We find that our sparse dictionaries improve the Pareto frontier of edit magnitude vs thoroughness of editing. In both figures, the feature dictionaries were trained on the first 10,000 elements of the Pile \citep{gao2020pile} (approximately 7 million activations) using the indicated $\alpha$ and $R$ values, on layer 11 of Pythia-410M (see Appendix \ref{appendix:ioi_other_layers} for results on other layers).}

    \label{fig:ioi-feature-identification}
\end{figure}

In this section, we quantify whether our learned dictionary features localise a specific model behaviour more tightly than the PCA decomposition of the model's activations. We do this via activation patching, a form of causal mediation analysis \citep{vig2020investigating}, through which we edit the model's internal activations along the directions indicated by our dictionary features and measure the changes to the model's outputs. We find that our dictionary features require fewer patches to reach a given level of KL divergence on the task studied than comparable decompositions (Figure \ref{fig:ioi-feature-identification}).

Specifically, we study model behaviour on the Indirect Object Identification (IOI) task \citep{wang2022interpretability}, in which the model completes sentences like ``Then, Alice and Bob went to the store. Alice gave a snack to \underline{\hspace{5mm}}''. This task was chosen because it captures a simple, previously-studied model behaviour. Recall that the training of our feature dictionaries does not emphasize any particular task.

\subsection{Adapting activation patching to dictionary features}
\label{subsec:activation patching}

In our experiment, we run the model on a counterfactual target sentence, which is a variant of the base IOI sentence with the indirect object changed (e.g., with ``Bob'' replaced by ``Vanessa''); save the encoded activations of our dictionary features; and use the saved activations to edit the model's residual stream when run on the base sentence.

In particular, we perform the following procedure. Fix a layer of the model to intervene on. Run the model on the target sentence, saving the model output logits $\mathbf{y}$ and the encoded features $\bar{\mathbf{c}}_1, ..., \bar{\mathbf{c}}_k$ of that layer at each of the $k$ tokens. Then, run the model on the base sentence up through the intervention layer, compute the encoded features $\mathbf{c}_1, ..., \mathbf{c}_k$  at each token, and at each position replace the residual stream vector $\mathbf{x}_i$ with the patched vector

\[ \mathbf{x}'_i = \mathbf{x}_i + \displaystyle \sum_{j \in F} (\bar{\mathbf{c}}_{i,j} -\mathbf{c}_{i,j})\mathbf{f}_j\]

where $F$ is the subset of the features which we intervene on (we describe the selection process for $F$ later in this section). Let $\mathbf{z}$ denote the output logits of the model when you finish applying it to the patched residual stream $\mathbf{x}'_1,...,\mathbf{x}'_k$. Finally, compute the KL divergence $D_{KL}(\mathbf{z}||\mathbf{y})$, which measures how close the patched model's predictions are to the target's. We compare these interventions to equivalent interventions using principal components found as in Section \ref{subsection:interpretability_baselines}.

To select the feature subset $F$, we use the Automated Circuit Discovery (ACDC) algorithm of \citet{conmy2023towards}. In particular, we use their Algorithm 4.1 on our features, treating them as a flat computational graph in which every feature contributes an independent change to the $D_{KL}$ output metric, as described above and averaged over a test set of 50 IOI data points. The result is an ordering on the features so that patching the next feature usually results in a smaller $D_{KL}$ loss than each previous feature. Then our feature subsets $F$ are the first $k$ features under this ordering. We applied ACDC separately on each decomposition.

\subsection{Precise Localisation of IOI Dictionary Features}

We show in Figure \ref{fig:ioi-feature-identification} that our sparse feature dictionaries allow the same amount of model editing, as measured by KL divergence from the target, in fewer patches (Left) and with smaller edit magnitude (Right) than the PCA decomposition. We also show that this does not happen if we train a non-sparse dictionary ($\alpha=0$). However, dictionaries with a larger sparsity coefficient $\alpha$ have lower overall reconstruction accuracy which appears in Figure \ref{fig:ioi-feature-identification} as a larger minimum KL divergence. In Figure \ref{fig:ioi-feature-identification} we consider interventions on layer 11 of the residual stream, and we plot interventions on other layers in Appendix \ref{appendix:ioi_other_layers}.

\section{Case Studies}

\label{section:case_studies}

\label{section:monosemantic}
In this section, we investigate individual dictionary features, highlighting several that appear to correspond to a single human-understandable explanation (i.e., that are monosemantic). We perform three analyses of our dictionary features to determine their semantic meanings: (1) \textit{Input}: We identify which tokens activate the dictionary feature and in which contexts, (2) \textit{Output}: We determine how ablating the feature changes the output logits of the model, and (3) \textit{Intermediate features}: We identify the dictionary features in previous layers that cause the analysed feature to activate.

\subsection{Input: Dictionary Features are Highly Monosemantic}
We first analyse our dictionary directions by checking what text causes them to activate. An idealised monosemantic dictionary feature will only activate on text corresponding to a single real-world feature, whereas a polysemantic dictionary feature might activate in unrelated contexts.

\begin{figure}[ht]
    \centering
    \begin{minipage}[b]{0.45\textwidth}
        \includegraphics[width=.9\textwidth]{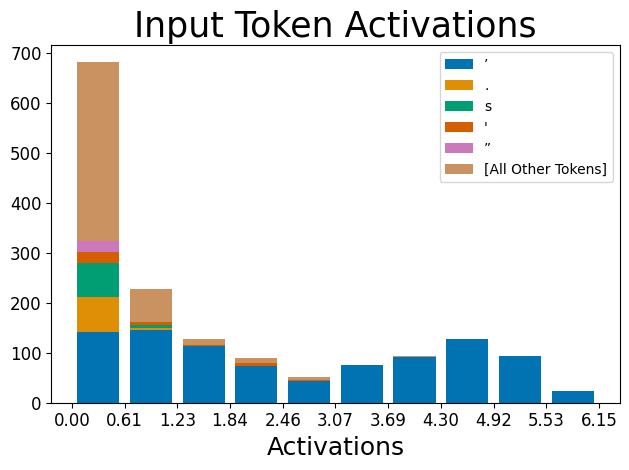}
    \end{minipage}
    \hfill  
    \begin{minipage}[b]{0.45\textwidth}
        \includegraphics[width=.9\textwidth]{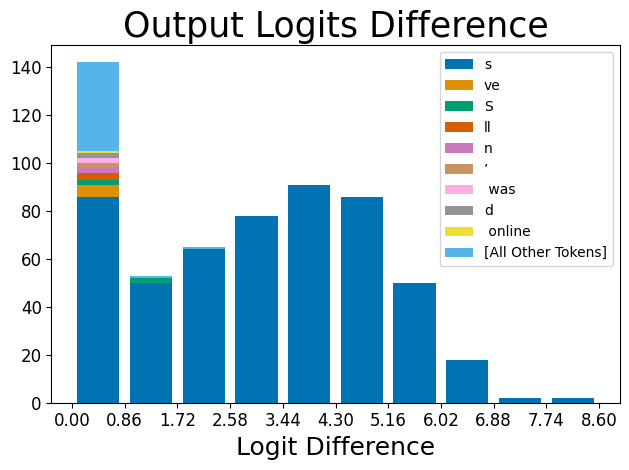}
    \end{minipage}
    \caption{Histogram of token counts for dictionary feature 556. (Left) For all datapoints that activate dictionary feature 556, we show the count of each token in each activation range. The majority of activations are apostrophes, particularly for higher activations. Notably the lower activating tokens are conceptually similar to apostrophes, such as other punctuation. (Right) We show which token predictions are suppressed by ablating the feature, as measured by the difference in logits between the ablated and unablated model. We find that the token whose prediction decreases the most is the ``s'' token. Note that there are 12k logits negatively effected, but we set a threshold of 0.1 for visual clarity.}
    \label{fig:apostrophe}
\end{figure}

To better illustrate the monosemanticity of certain dictionary features, we plot the histogram of activations across token activations. This technique only works for dictionary features that activate for a small set of tokens. We find dictionary features that only activate on apostrophes (Figure \ref{fig:apostrophe}); periods; the token `` the''; and newline characters. The apostrophe feature in Figure \ref{fig:apostrophe} stands in contrast to the default basis for the residual stream, where the dimension that most represents an apostrophe is displayed in Figure \ref{fig:neuron_basis_apostrophe} in Appendix \ref{appendix:residual_stream_basis}; this dimension is polysemantic since it represents different information at different activation ranges.

Although the dictionary feature discussed in the previous section activates only for apostrophes, it does not activate on \emph{all} apostrophes. This can be seen in Figures \ref{fig:ill_feature} and \ref{fig:dont_feature} in Appendix \ref{appendix:learned_features_examples}, showing two other apostrophe-activating dictionary features, but for different contexts (such as ``[I/We/They]'ll" and ``[don/won/wouldn]'t"). Details for how we searched and selected for dictionary features can be found in Appendix \ref{appendix:feature_search_details}.

\subsection{Output: Dictionary Features have Intuitive Effects on the Logits}
In addition to looking at which tokens activate the dictionary feature, we investigate how dictionary features affect the model's output predictions for the next token by ablating the feature from the residual stream\footnote{Specifically we use less-than-rank-one ablation, where we lower the activation vector in the direction of the feature only up to the point where the feature is no longer active.}. If our dictionary feature is interpretable, subtracting its value from the residual stream should have a logical effect on the predictions of the next token. We see in Figure \ref{fig:apostrophe} (Right) that the effect of removing the apostrophe feature mainly reduces the logit for the following ``s". This matches what one would expect from a dictionary feature that detects apostrophes and is used by the model to predict the ``s'' token that would appear immediately after the apostrophe in possessives and contractions like ``let's''. 

\subsection{Intermediate Features: Dictionary Features Allow Automatic Circuit Detection}
We can also understand dictionary features in relation to the upstream and downstream dictionary features: given a dictionary feature, which dictionary features in previous layers cause it to activate, and which dictionary features in later layers does it cause to activate?

\begin{figure}[ht]
    \centering
    \includegraphics[width=\textwidth]{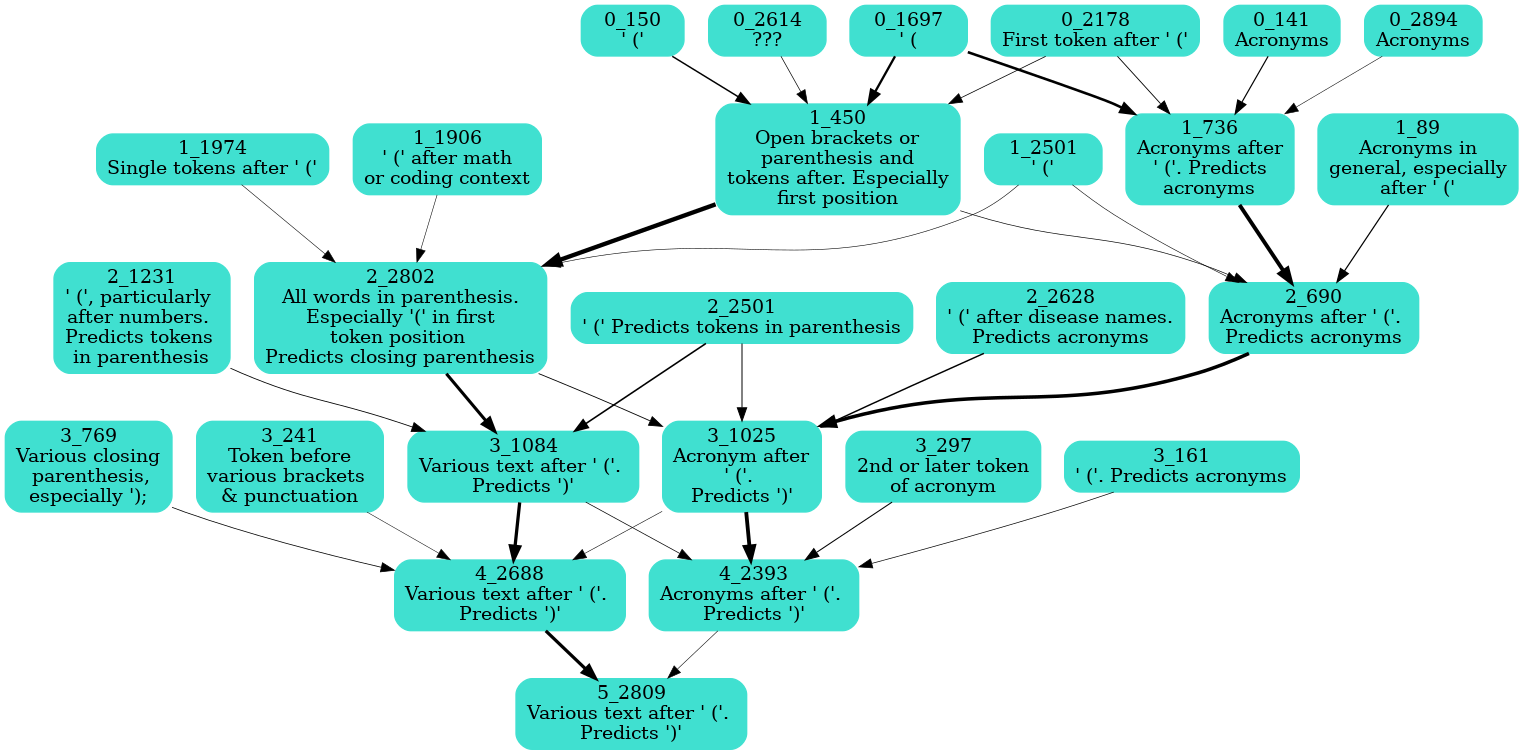}
    \caption{Circuit for the closing parenthesis dictionary feature, with human interpretations of each feature shown. Edge thickness indicates the strength of the causal effect between dictionary features in successive residual stream layers, as measured by ablations. Many dictionary features across layers correspond to similar real-world features and often point in similar directions in activation space, as measured by cosine similarity.}
    \label{fig:closing_parentheses_circuit}
\end{figure}

To automatically detect the relevant dictionary features, we choose a target dictionary feature such as layer 5's feature for tokens in parentheses which predicts a closing parentheses (Figure \ref{fig:closing_parentheses_circuit}). For this target dictionary feature, we find its maximum activation $M$ across our dataset, then sample 20 contexts that cause the target feature to activate in the range $[M/2, M]$. For each dictionary feature in the previous layer, we rerun the model while ablating this feature and sort the previous-layer features by how much their ablation decreased the target feature. If desired, we can then recursively apply this technique to the dictionary features in the previous layer with a large impact. The results of this process form a causal tree, such as Figure \ref{fig:closing_parentheses_circuit}. 

Being the last layer, layer 5's role is to output directions that directly correspond to tokens in the unembedding matrix. In fact, when we unembed feature $5_{2027}$, the top-tokens are all closing parentheses variations. Intuitively, previous layers will detect all situations that precede closing parentheses, such as dates, acronyms, and phrases.


\section{Discussion}
\subsection{Related Work}

A limited number of previous works have learned dictionaries of sparsely-activating features in pre-trained models, including \citet{yun2021transformer} and \citet{sharkey2023technical}, the latter of which motivated this work. However, similar methods have been applied in other domains, in particular in understanding neurons in the visual cortex \citep{olshausen2004sparse, wright2022high}. 

In contrast to our approach, where we try to impose sparsity after training, many previous works have encouraged sparsity in neural networks via changes to the architecture or training process. These approaches include altering the attention mechanism \citep{correia2019adaptively}, adding $\ell^1$ penalties to neuron activations \citep{kasioumis2021elite, georgiadis2019accelerating}, pruning neurons \citep{frankle2018lottery}, and using the softmax function as the non-linearity in the MLP layers \citep{elhage2022solu}. However, training a state-of-the-art foundation model with these additional constraints is difficult \citep{elhage2022solu}, and improvements to interpretability are not always realized \citep{Meister2021IsSA}.

\subsection{Limitations and Future Work}

While we have presented evidence that our dictionary features are interpretable and causally important, we do not achieve 0 reconstruction loss (Equation \ref{eqn:loss_function}), indicating that our dictionaries fail to capture all the information in a layer's activations. We have also confirmed this by measuring the perplexity of the model's predictions when a layer is substituted with its reconstruction. For instance, replacing the residual stream activations in layer 2 of Pythia-70M with our reconstruction of those activations increases the perplexity on the Pile \citep{gao2020pile} from 25 to 40. To reduce this loss of information, we would like to explore other sparse autoencoder architectures and to try minimizing the change in model outputs when replacing the activations with our reconstructed vectors, rather than the reconstruction loss. Future efforts could also try to improve feature dictionary discovery by incorporating information about the weights of the model or dictionary features found in adjacent layers into the training process.

Our current methods for training sparse autoencoders are best suited to the residual stream. There is evidence that they may be applicable to the MLPs (see Appendix \ref{appendix:full_autointerp}), but the training pipeline used to train the dictionaries in this paper is not able to robustly learn overcomplete bases in the intermediate layers of the MLP. We're excited by future work investigating what changes can be made to better understand the computations performed by the attention heads and MLP layers, each of which poses different challenges.

In Section \ref{section:identifying_causal_features}, we show that for the IOI task, behaviour is dependent on a relatively small number of features. Because our dictionary is trained in a task-agnostic way, we expect this result to generalize to similar tasks and behaviours, but more work is needed to confirm this suspicion. If this property generalizes, we would have a set of features which allow for understanding many model behaviours using just a few features per behaviour. We would also like to trace the causal dependencies between features in different layers, with the overarching goal of providing a lens for viewing language models under which causal dependencies are sparse. This would hopefully be a step towards the eventual goal of building an end-to-end understanding of how a model computes its outputs.

\subsection{Conclusion}

Sparse autoencoders are a scalable, unsupervised approach to disentangling language model network features from superposition. Our approach requires only unlabelled model activations and uses orders of magnitude less compute than the training of the original models. We have demonstrated that the dictionary features we learn are more interpretable by autointerpretation, letting us pinpoint the features responsible for a given behaviour more finely, and are more monosemantic than comparable methods. This approach could facilitate the mapping of model circuits, targeted model editing, and a better understanding of model representations.

An ambitious dream in the field of interpretability is enumerative safety \citep{elhage2022toy}: producing a human-understandable explanation of a model's computations in terms of a complete list of the model's features and thereby providing a guarantee that the model will not perform dangerous behaviours such as deception. We hope that the techniques we presented in this paper also provide a step towards achieving this ambition.

\ificlrfinal
\subsubsection*{Acknowledgments}
We would like to thank the OpenAI Researcher Access Program for their grant of model credits for the autointerpretation and CoreWeave for providing EleutherAI with the computing resources for this project. We also thank Nora Belrose, Arthur Conmy, Jake Mendel, and the OpenAI Automated Interpretability Team (Jeff Wu, William Saunders, Steven Bills, Henk Tillman,
and Daniel Mossing) for valuable discussions regarding the design of various experiments. We thank Wes Gurnee, Adam Jermyn, Stella Biderman, Leo Gao, Curtis Huebner, Scott Emmons, and William Saunders for their feedback on earlier versions of this paper. Thanks to Delta Hessler for proofreading. LR is supported by the Long Term Future Fund. RH is supported by an Open Philanthropy grant. HC was greatly helped by the MATS program, funded by AI Safety Support.

\fi

\bibliography{iclr2024_conference_google_scholar_version}
\bibliographystyle{iclr2024_conference}

\appendix

\section{Autointerpretation Protocol}
\label{appendix:autointerp_protocol}
The autointerpretability process consists of five steps and yields both an interpretation and an autointerpretability score:

\begin{enumerate}
    \item On each of the first 50,000 lines of OpenWebText, take a 64-token sentence-fragment, and measure the feature's activation on each token of this fragment. Feature activations are rescaled to integer values between 0 and 10.
    \item Take the 20 fragments with the top activation scores and pass 5 of these to GPT-4, along with the rescaled per-token activations. Instruct GPT-4 to suggest an explanation for when the feature (or neuron) fires, resulting in an interpretation.
    \item Use GPT-3.5\footnote{While the process described in \citet{bills2023language} uses GPT-4 for the simulation step, we use GPT-3.5. This is because the simulation protocol requires the model's logprobs for scoring, and OpenAI's public API for GPT-3.5 (but not GPT-4) supports returning logprobs.} to simulate the feature across another 5 highly activating fragments and 5 randomly selected fragments (with non-zero variation) by asking it to provide the per-token activations.
    \item Compute the correlation of the simulated activations and the actual activations. This correlation is the autointerpretability score of the feature. The texts chosen for scoring a feature can be random text fragments, fragments chosen for containing a particularly high activation of that feature, or an even mixture of the two. We use a mixture of the two unless otherwise noted, also called `top-random' scoring.
    \item If, amongst the 50,000 fragments, there are fewer than 20 which contain non-zero variation in activation, then the feature is skipped entirely.
\end{enumerate}

Although the use of random fragments in Step 4 is ultimately preferable given a large enough sample size, the small sample sizes of a total of 640 tokens used for analysis mean that a random sample will likely not contain any highly activating examples for all but the most common features, making top-random scoring a desirable alternative.

\section{Sparse Autoencoder Training and Hyperparameter Selection}
\label{appendix:hparam_selection}
To train the sparse autoencoder described in Section \ref{section:dictionary_learning}, we use data from the Pile \citep{gao2020pile}, a large, public webtext corpus. We run the model that we want to interpret over this text while caching and saving the activations at a particular layer. These activations then form a dataset, which we use to train the autoencoders. The autoencoders are trained with the Adam optimiser with a learning rate of 1e-3 and are trained on 5-50M activation vectors for 1-3 epochs, with larger dictionaries taking longer to converge. A single training run using this quantity of data completes in under an hour on a single A40 GPU.

\begin{figure}[hb]
    \centering
    \includegraphics[width=.9\textwidth]
    {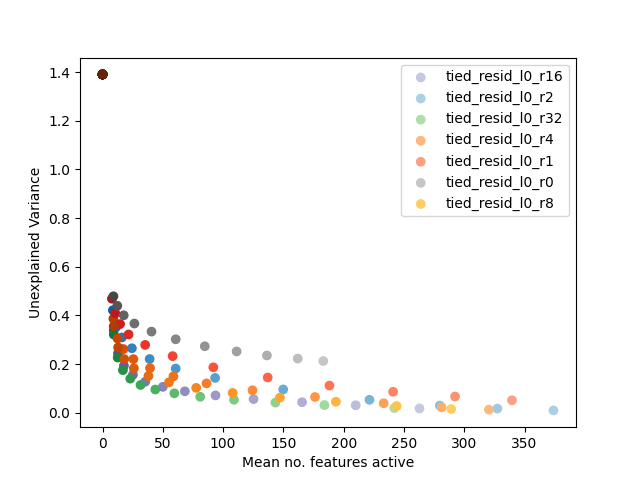}
    \caption{The tradeoff between the average number of features active and the proportion of variance that is unexplained for the MLP at layer 0.}
    \label{fig:smooth_trade}
\end{figure}

\begin{figure}
    \centering
    \includegraphics[width=0.9\textwidth]{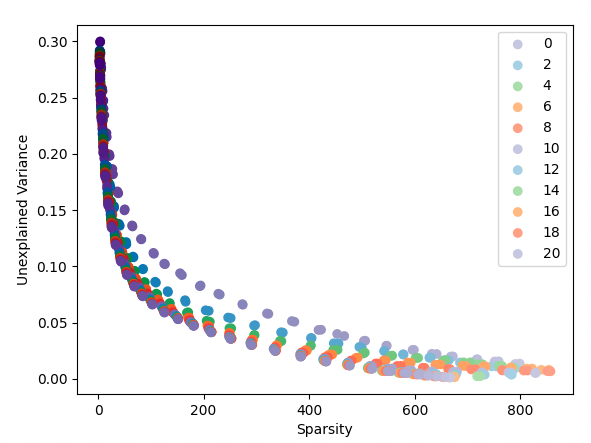}
    \caption{The tradeoff between sparsity and unexplained variance in our reconstruction. Each series of points is a sweep of the $\alpha$ hyperparameter, trained for the number of epochs given in the legend.}
    \label{fig:trade_over_time}
\end{figure}

When varying the hyperparameter $\alpha$ which controls the importance of the sparsity loss term, we consistently find a smooth tradeoff between the sparsity and accuracy of our autoencoder, as shown in Figure \ref{fig:smooth_trade}. The lack of a `bump' or `knee' in these plots provides some evidence that there is not a single correct way to decompose activation spaces into a sparse basis, though to confirm this would require many additional experiments. Figure \ref{fig:trade_over_time} shows the convergence behaviour of a set of models with varying $\alpha$ over multiple epochs.
 
\section{Further Autointerpretation Results}
\label{appendix:full_autointerp}
\subsection{Interpretability is Consistent across Dictionary Sizes}

\begin{figure}
    \centering
    \includegraphics[width=.8\textwidth]
    {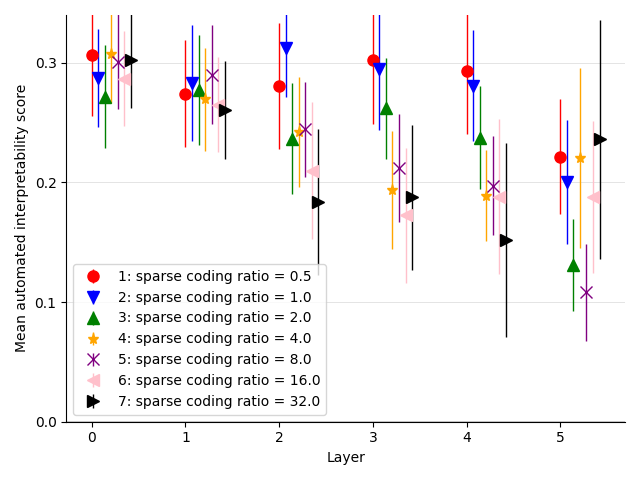}
    \caption{Comparison of average interpretability scores across dictionary sizes. All dictionaries were trained on 20M activation vectors obtained by running Pythia-70M over the Pile with $\alpha=.00086$.}
    \label{fig:dictionary_size_comparison}
\end{figure}

We find that larger interpretability scores of our feature dictionaries are not limited to overcomplete dictionaries (where the ratio, $R$, of dictionary features to model dimensions is $>1$), but occurs even in dictionaries that are smaller than the underlying basis, as shown in Figure \ref{fig:dictionary_size_comparison}. These small dictionaries are able to reconstruct the activation vectors less accurately, so with each feature being similarly interpretable, the larger dictionaries will be able to explain more of the overall variance.

\subsection{High Interpretability Scores Are Not an Artefact of Top Scoring}

A possible concern is that the autointerpretability method described in Section \ref{section:autointerp} combines top activating fragments (which are usually large) with random activations (which are usually small), making it relatively easy to identify activations. Following the lead of \citet{bills2023language}, we control for this by recomputing the autointerpretation scores by modifying Step 3 using only randomly selected fragments. With large sample sizes, using random fragments should be the true test of our ability to interpret a potential feature. However, the features we are considering are heavy-tailed, so with limited sample sizes, we should expect random samples to underestimate the true correlation.

\begin{figure}
    \centering
    \includegraphics[width=0.8\textwidth]    {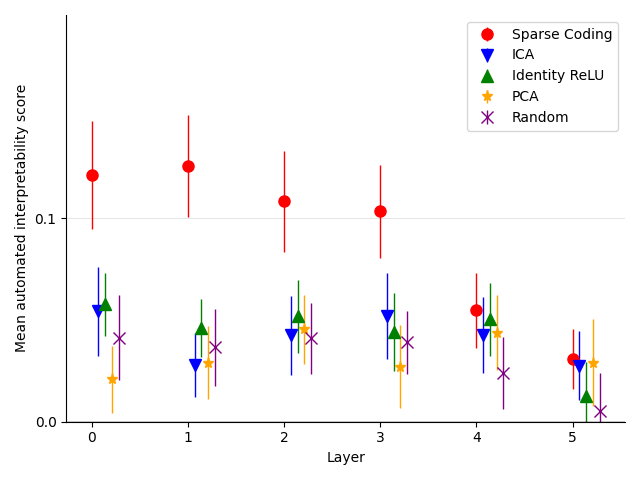}
    \caption{Random-only interpretability scores across each layer, a measure of how well the interpretation of the top activating cluster is able to explain the entire range of activations.}
    \label{fig:residual_random_activations}
\end{figure}

In Figure \ref{fig:residual_random_activations} we show autointerpretability scores for fragments using only random fragments. Matching \citet{bills2023language}, we find that random-only scores are significantly smaller than top-and-random scores, but also that our learned features still consistently outperform the baselines, especially in the early layers. Since our learned features are more sparse than the baselines and thus, activate less on a given fragment, this is likely to underestimate the performance of sparse coding relative to baselines.

An additional potential concern is that the structure of the autoencoders allows them to be sensitive to less than a full direction in the activation space, resulting in an unfair comparison. We show in Appendix \ref{appendix:topk} that this is not the source of the improved performance of sparse coding.


While the residual stream can usually be treated as a vector space with no privileged basis (a basis in which we would expect changes to be unusually meaningful, such as the standard basis after a non-linearity in an MLP), it has been noted that there is a tendency for transformers to store information in the residual stream basis\citep{dettmers2022llm}, which is believed to be caused by the Adam optimiser saving gradients with finite precision in the residual basis\citep{elhage2023privileged}. We do not find residual stream basis directions to be any more interpretable than random directions.

\subsection{Interpreting the MLP Sublayer}

Our approach of learning a feature dictionary and interpreting the resulting features can, in principle, be applied to any set of internal activations of a language model, not just the residual stream. Applying our approach to the MLP sublayer of a transformer resulted in mixed success. Our approach still finds many features that are more interpretable than the neurons. However, our architecture also learns many dead features, which never activate across the entire corpus. In some cases, there are so many dead features that the set of living features does not form an overcomplete basis. For example, in a dictionary with twice as many features as neurons, less than half might be active enough to perform automatic interpretability. The exceptions to this are the early layers, where a large fraction of them are active. 

For learning features in MLP layers, we find that we retain a larger number of features if we use a different matrix for the encoder and decoder, so that Equations \ref{eq:encoder} and \ref{eq:decoder} become

\begin{eqnarray}
\mathbf{c} &=& ReLU(M_e\mathbf{x}+\mathbf{b}) \label{eq:untied_encoder}\\
\mathbf{\hat x} &=& M_d^T \mathbf{c} \label{eq:untied_decoder}
\end{eqnarray}

\begin{figure}
    \label{fig:topk_mlp}
    \centering
    \includegraphics[width=.9\textwidth]{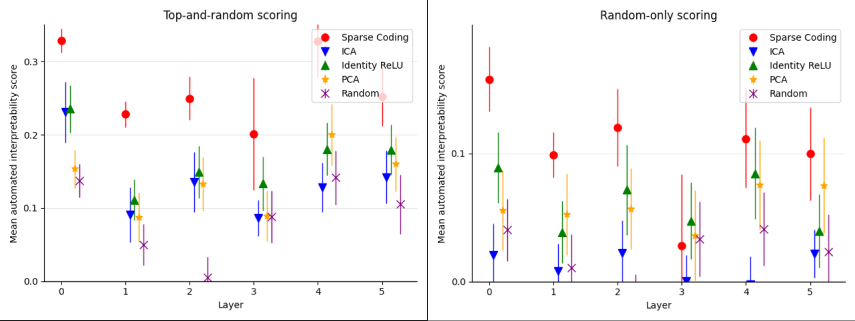}
    \caption{Top-and-random and random-only interpretability scores for across each MLP layer, using an $\ell^1$ coefficient $\alpha=3.2e-4$ and dictionary size ratio $R=1$.}
\end{figure}

We are currently working on methods to overcome this and find truly overcomplete bases in the middle and later MLP layers.

\subsection{Interpretability Scores Correlate with Kurtosis and Skew of Activation}

It has been shown that the search for sparse, overcomplete dictionaries can be reformulated in terms of the search for directions that maximise the $\ell^4$-norm \citep{qu2019analysis}.

We offer a test of the utility of this by analysing the correlation between interpretability and a number of properties of learned directions. We find that there is a correlation of 0.19 and 0.24 between the degree of positive skew and kurtosis respectively that feature activations have and their top-and-random interpretability scores, as shown in Table \ref{tab:moment_correlation}.

\begin{table}[h]
    \centering
    \begin{tabular}{|c|c|}
        \hline
        \textbf{Moment} & \textbf{Correlation with top-random interpretability score} \\
        \hline
        Mean & -0.09\\
        \hline
        Variance & 0.02 \\
        \hline
        Skew & 0.20 \\
        \hline
        Kurtosis & 0.15 \\
        \hline
    \end{tabular}
    \caption{Correlation of interpretability score with feature moments across residual stream results, all layers, with dictionary size ratios $R \in \{0.5, 1, 2, 4, 8\}$.}
    \label{tab:moment_correlation}
\end{table}

This also accords with the intuitive explanation that the degree of interference due to other active features will be roughly normally distributed by the central limit theorem. If this is the case, then features will be notable for their heavy-tailedness.

This also explains why Independent Component Analysis (ICA), which maximises the non-Gaussianity of the found components, is the best performing of the alternatives that we considered.

\section{Qualitative Feature Analysis}
\label{appendix:feature_interp_examples}
\subsection{Residual Stream Basis}
\label{appendix:residual_stream_basis}

\begin{figure}[ht]
    \centering
    \includegraphics[width=0.8\textwidth]{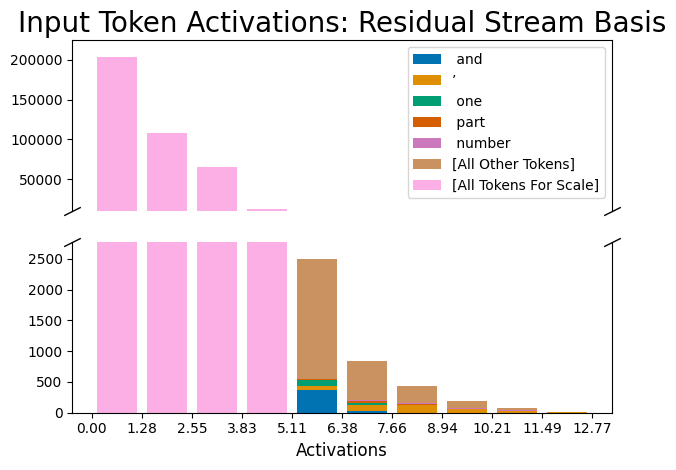}
    \caption{Histogram of token counts in the neuron basis. Although there are a large fraction of apostrophes in the upper activation range, this only explains a very small fraction of the variance for middle-to-lower activation ranges.}
    \label{fig:neuron_basis_apostrophe}
\end{figure}

Figure \ref{fig:neuron_basis_apostrophe} gives a token activation histogram of the residual stream basis. Connecting this residual stream dimension to the apostrophe feature from Figure \ref{fig:apostrophe}, this residual dimension was the 10th highest dimension read from the residual stream by our feature\footnote{The first 9 did not have apostrophes in their top-activations like dimension 21.}. 

\subsection{Examples of Learned Features}
\label{appendix:learned_features_examples}

\begin{figure}[ht]
    \centering
    \begin{minipage}[b]{0.45\textwidth}
        \includegraphics[width=\textwidth]{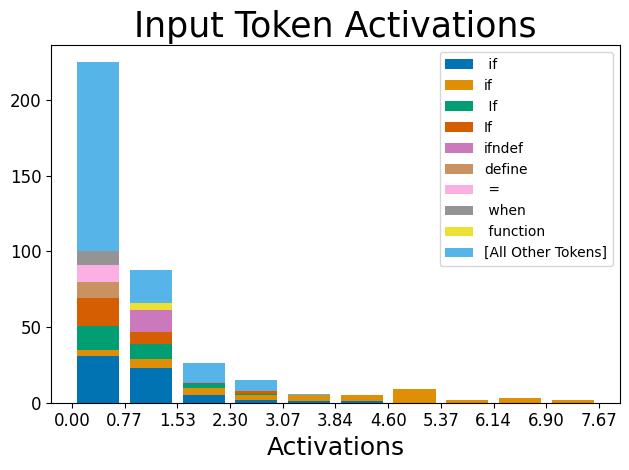}
    \end{minipage}
    \hfill  
    \begin{minipage}[b]{0.45\textwidth}
        \includegraphics[width=\textwidth]{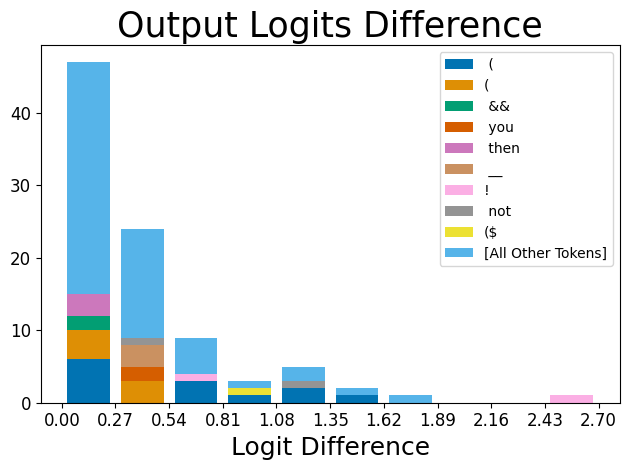}
    \end{minipage}
    \caption{`If' feature in coding contexts}
    \label{fig:if_feature}
\end{figure}

\begin{figure}[ht]
    \centering
    \begin{minipage}[b]{0.49\textwidth}
        \includegraphics[width=0.9\textwidth]{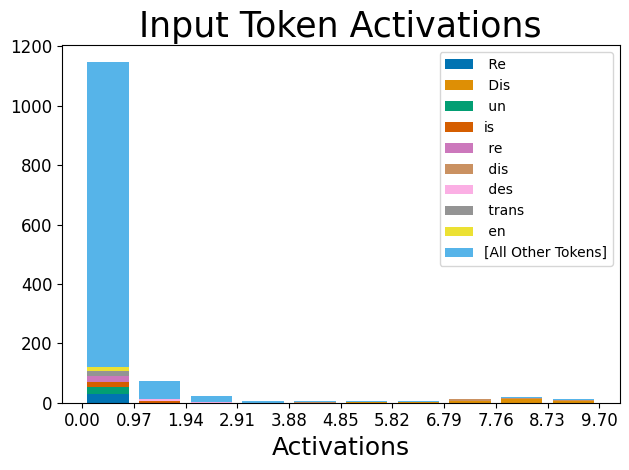}
    \end{minipage}
    \hfill  
    \begin{minipage}[b]{0.49\textwidth}
        \includegraphics[width=0.9\textwidth]{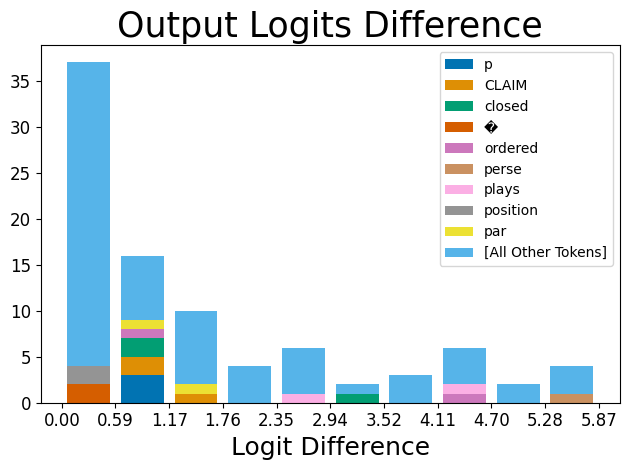}
    \end{minipage}
    \caption{`Dis' token-level feature showing bigrams, such as `disCLAIM', `disclosed', `disordered', etc.}
    \label{fig:dis_feature}    
\end{figure}

\begin{figure}[ht]
    \centering
    \begin{minipage}[b]{0.49\textwidth}
        \includegraphics[width=0.9\textwidth]{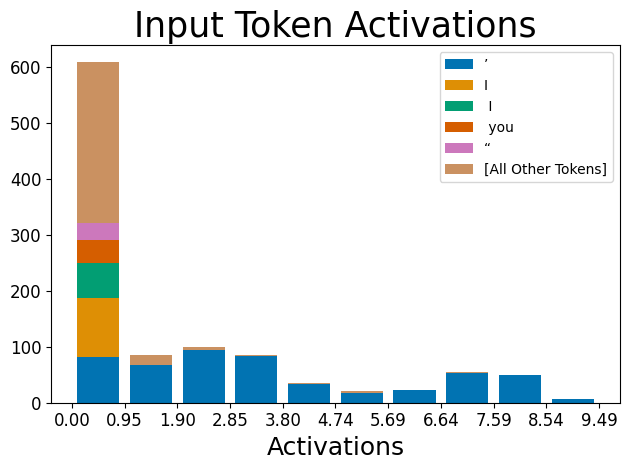}
    \end{minipage}
    \hfill  
    \begin{minipage}[b]{0.49\textwidth}
        \includegraphics[width=0.9\textwidth]{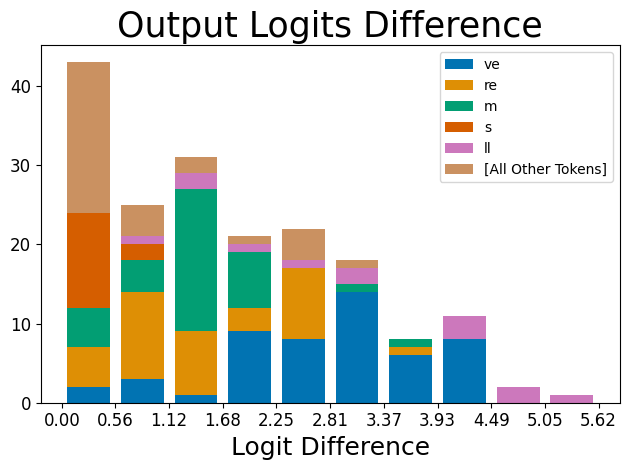}
    \end{minipage}
    \caption{Apostrophe feature in ``I'll"-like contexts.}
    \label{fig:ill_feature}
\end{figure}

\begin{figure}[ht]
    \centering
    \begin{minipage}[b]{0.49\textwidth}
        \includegraphics[width=0.9\textwidth]{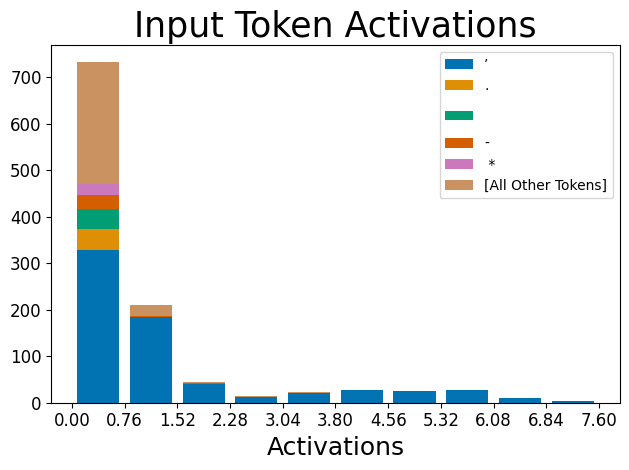}
    \end{minipage}
    \hfill  
    \begin{minipage}[b]{0.49\textwidth}
        \includegraphics[width=0.9\textwidth]{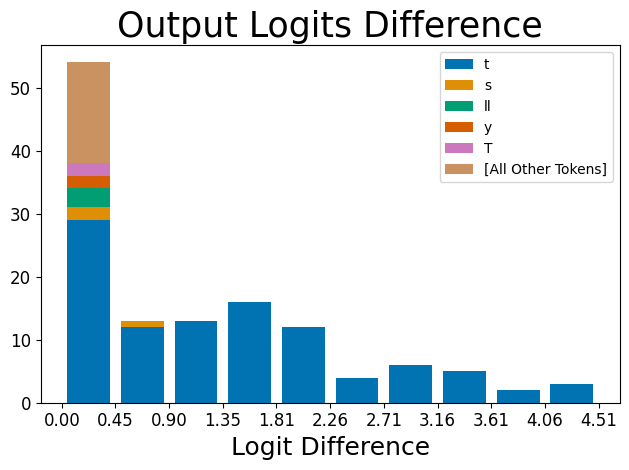}
    \end{minipage}
    \caption{Apostrophe feature in ``don't"-like contexts.}
    \label{fig:dont_feature}
\end{figure}

Other features are shown in Figures \ref{fig:if_feature}, \ref{fig:dis_feature}, \ref{fig:ill_feature}, and \ref{fig:dont_feature}.

\subsection{Feature Search Details}
\label{appendix:feature_search_details}

We searched for the apostrophe feature using the sentence `` I don't know about that. It is now up to Dave''', and seeing which feature (or residual stream dimension) activates the most for the last apostrophe token. The top activating feature in our dictionary was an outlier dimension feature (i.e., a feature direction that mainly reads from an outlier dimension of the residual stream), the apostrophes after O (and predicted O'Brien, O'Donnell, O'Connor, O'clock, etc), then the apostrophe-preceding-s feature. 

For the residual basis dimension, we searched for max and min activating dimensions (since the residual stream can be both positive and negative), where the top two most positive dimensions were outlier dimensions, the top two negative dimensions were our displayed one and another outlier dimension, respectively.

\subsection{Failed Interpretability Methods}
\label{appendix:failed_interp_methods}

We attempted a weight-based method going from the dictionary in layer 4 to the dictionary in layer 5 by multiplying a feature by the MLP and checking the cosine similarity with features in layer 5. There were no meaningful connections. Additionally, it's unclear how to apply this to the Attention sublayer since we'd need to see which position dimension the feature is in. We expected this failed by going out of distribution.

\section{Number of Active Features}
\label{appendix:n_features_active}
\begin{figure}[h]
    \centering
    \includegraphics[width=0.9\textwidth]{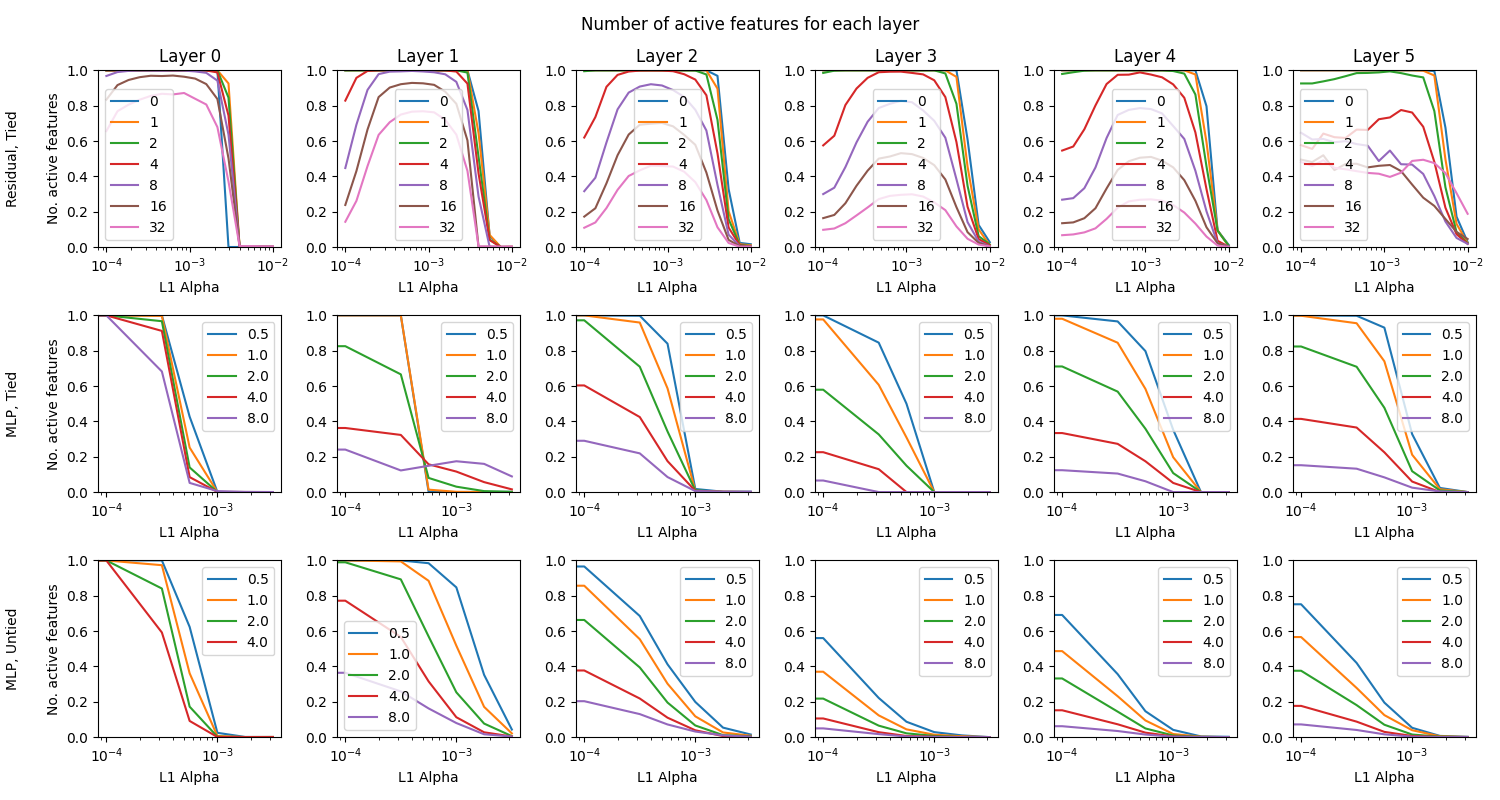}
    \caption{The number of features that are active, defined as activating more than 10 times across 10M datapoints, changes with sparsity hyperparamter $\alpha$ and dictionary size ratio $R$.}
    \label{fig:n_active_summary}
\end{figure}

In Figure \ref{fig:n_active_summary} we see that, for residual streams, we consistently learn dictionaries that are at least 4x overcomplete before some features start to drop out completely, with the correct hyperparameters. For MLP layers you see large numbers of dead features even with hyperparameter $\alpha=0$. These figures informed the selection of $\alpha=8.6e-4$ and $\alpha=3.2e-4$ that went into the graphs in Section \ref{section:autointerp} for the residual stream and MLP respectively. Due to the large part of the input space that is never used due to the non-linearity, it is much easier for MLP dictionary features to become stuck at a position where they hardly ever activate. In future we plan to reinitialise such `dead features' to ensure that we learn as many useful dictionary features as possible.



\section{Editing IOI Behaviour on other Layers}
\label{appendix:ioi_other_layers}
\begin{figure}
     \centering
     \begin{subfigure}[b]{0.49\textwidth}
         \centering
         \includegraphics[width=\textwidth]{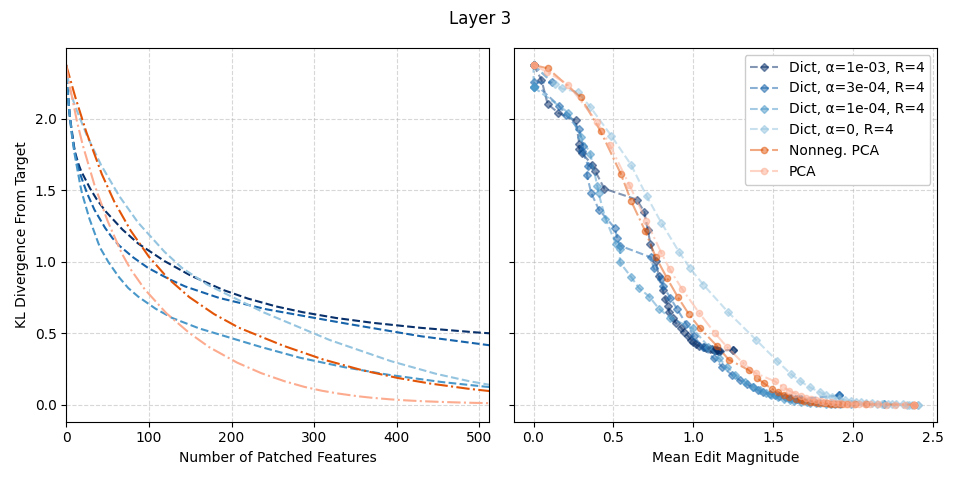}
     \end{subfigure}
     \hfill
     \begin{subfigure}[b]{0.49\textwidth}
         \centering
         \includegraphics[width=\textwidth]{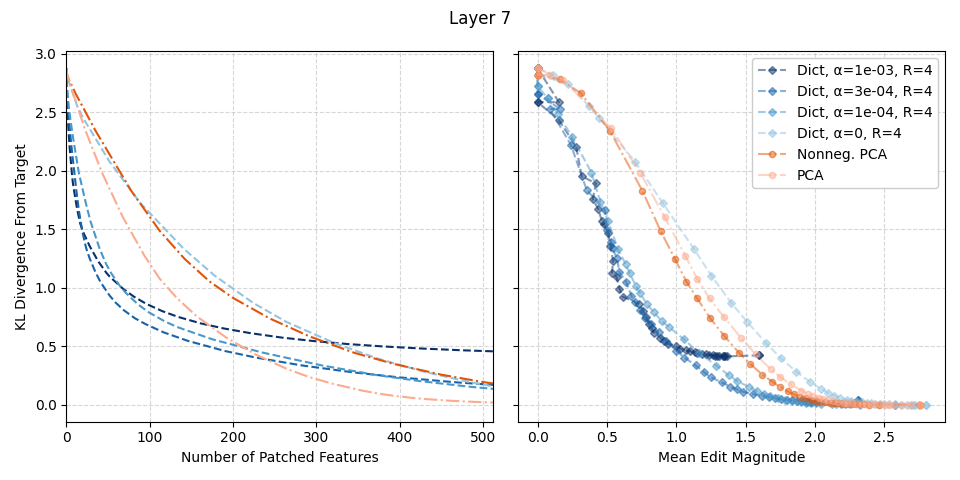}
     \end{subfigure} \\
     \begin{subfigure}[b]{0.49\textwidth}
         \centering
         \includegraphics[width=\textwidth]{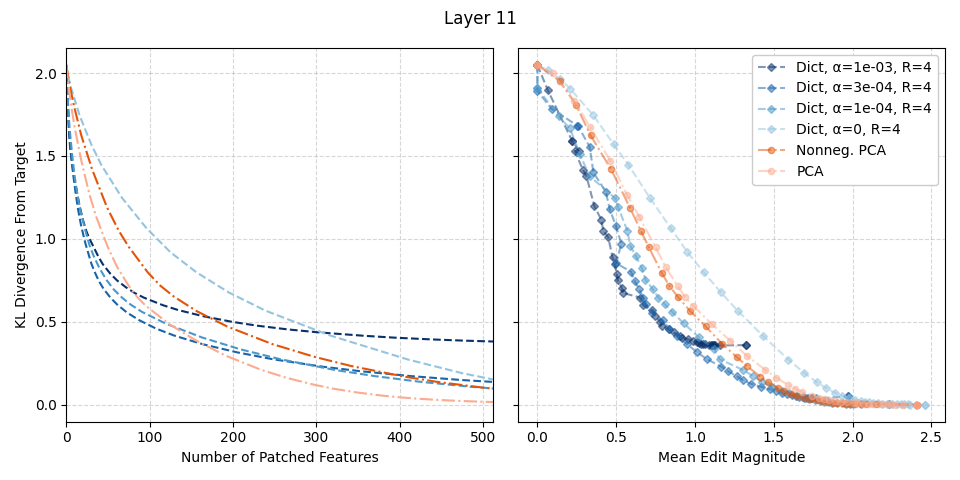}
     \end{subfigure}
     \hfill
     \begin{subfigure}[b]{0.49\textwidth}
         \centering
         \includegraphics[width=\textwidth]{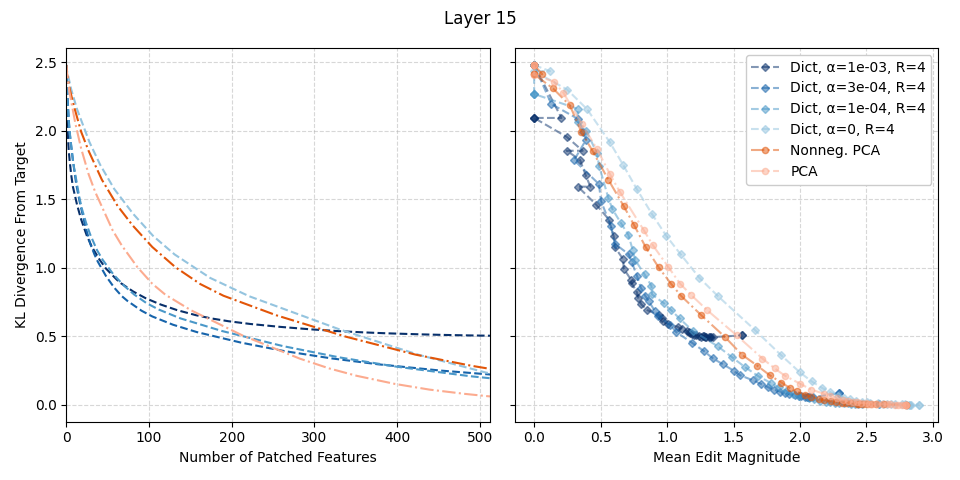}
     \end{subfigure}\\
     \begin{subfigure}[b]{0.49\textwidth}
         \centering
         \includegraphics[width=\textwidth]{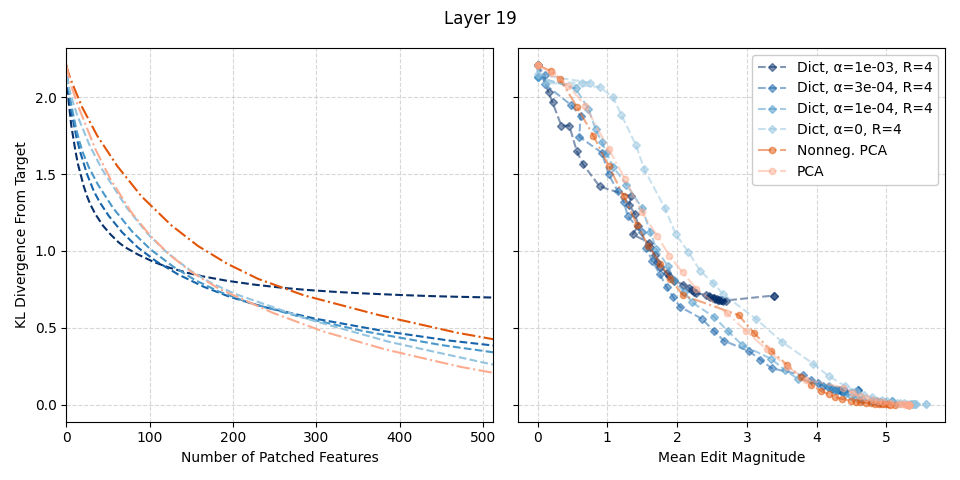}
     \end{subfigure}
     \hfill
     \begin{subfigure}[b]{0.49\textwidth}
         \centering
         \includegraphics[width=\textwidth]{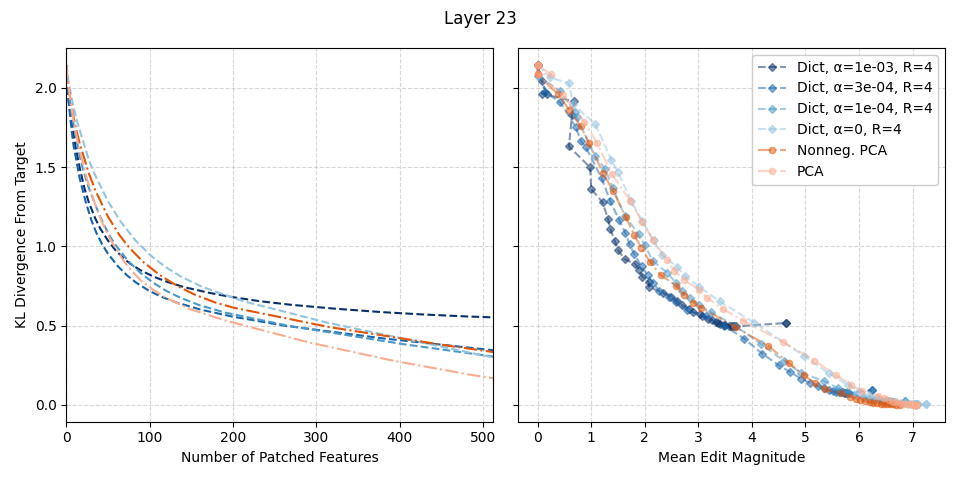}
     \end{subfigure}
        \caption{Divergence from target output against number of features patched and magnitude of edits for layers 3, 7, 11, 15, 19 and 23 of the residual stream of Pythia-410M. Pythia-410M has 24 layers, which we index 0, 1, ..., 23.}
        \label{fig:three graphs}

\end{figure}

In Figure \ref{fig:three graphs} we show results of the procedure in Section \ref{section:identifying_causal_features} across a range of layers in Pythia-410M.

\section{Top K Comparisons}
\label{appendix:topk}
\begin{figure}
    \centering
    \includegraphics[width=.7\textwidth]{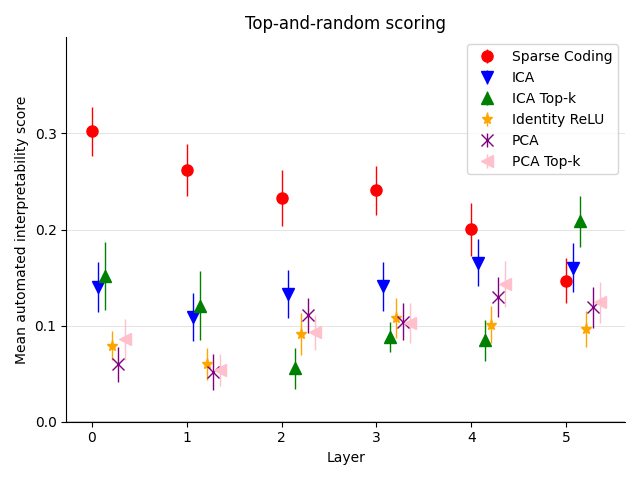}
    \caption{Autointerpretation scores across layers for the residual stream, including top-K baselines for ICA and PCA.}
    \label{fig:topk}
\end{figure}

As mentioned in Section \ref{section:autointerp}, the comparison directions learnt by sparse coding and those in the baselines are not perfectly even. This is because, for example, a PCA direction is active to an entire half-space on one side of a hyperplane through the origin, whereas  a sparse coding feature activates on less than a full direction, being only on the far side of a hyperplane that does not intersect the origin. This is due to the bias applied before the activation, which is, in practice, always negative. To test whether this difference is responsible for the higher scores, we run a variant of PCA and ICA in which we have a fixed number of directions, K,  which can be active for any single datapoint. We set this K to be equal to the average number of active features for a sparse coding dictionary with ratio $R=1$ and $\alpha=8.6e-4$ trained on the layer in question. We compare the results in Figure \ref{fig:topk}, showing that this change does not explain more than a small fraction of the improvement in scores.


\end{document}